\documentclass[9pt,conference]{IEEEtran}
\IEEEoverridecommandlockouts
\def\BibTeX{{\rm B\kern-.05em{\sc i\kern-.025em b}\kern-.08em
    T\kern-.1667em\lower.7ex\hbox{E}\kern-.125emX}}

\usepackage{comment}

\usepackage{float}
\usepackage{stfloats}
\usepackage{algorithm}
\usepackage{graphicx}
\usepackage[noend]{algpseudocode}
\usepackage{url}
\usepackage{breqn}
\usepackage{cite}
\usepackage[usenames]{color}
\usepackage{amsfonts}
\usepackage{fancyhdr}
\usepackage{todonotes}
\usepackage{booktabs}
\usepackage{arydshln}
\usepackage{xcolor}

\usepackage{amsmath,amssymb,amsthm}

\usepackage{psfrag, caption,subcaption}
\newtheorem{theorem}{Theorem}
\newtheorem{Proposition}{Proposition}
\newtheorem{lemma}{Lemma}

\usepackage{arydshln}

\usepackage{hyperref}
\usepackage{acronym}
\acrodef{gcn}[GCN]{graph convolutional network}
\acrodef{gnn}[GNN]{graph neural network}
\acrodef{grl}[GRL]{graph representation learning}
\acrodef{st}[ST]{graph diffusion scattering transform}
\acrodef{gat}[GAT]{graph attention transformer}
\acrodef{gft}[GFT]{graph Fourier transform}

\def\cast{{
   \mathord{
      \hbox to 0em{
         \ooalign{
	   \smash{\hbox{$\ast$}}\crcr
	   \smash{\hskip-1pt\Large\hbox{$\circ$}} }
	 \hidewidth}
      \phantom{\bigcirc}
} }}

\newcommand{\bds}{\begin {itemize}}
\newcommand{\eds}{\end {itemize}}
\newcommand{\bdf}{\begin{definition}}
\newcommand{\blm}{\begin{lemma}}
\newcommand{\edf}{\end{definition}}
\newcommand{\elm}{\end{lemma}}
\newcommand{\bthm}{\begin{theorem}}
\newcommand{\ethm}{\end{theorem}}
\newcommand{\bprp}{\begin{prop}}
\newcommand{\eprp}{\end{prop}}
\newcommand{\bcl}{\begin{claim}}
\newcommand{\ecl}{\end{claim}}
\newcommand{\bcr}{\begin{coro}}
\newcommand{\ecr}{\end{coro}}
\newcommand{\bquest}{\begin{question}}
\newcommand{\equest}{\end{question}}

\newcommand{\larrow}{{\larrow}}

\newcommand{\argmin}{\ensuremath{\mathrm{arg}\min}}
\newcommand{\argmax}{\ensuremath{\mathrm{arg}\max}}

\newcommand{\cE}{{\ensuremath{\mathcal{E}}}}

\newcommand{\cG}{{\ensuremath{\mathcal{G}}}}

\newcommand{\cL}{{\ensuremath{\mathcal{L}}}}

\newcommand{\cP}{{\ensuremath{\mathcal{P}}}}

\newcommand{\cV}{{\ensuremath{\mathcal{V}}}}

\newcommand{\vu}{{\ensuremath{{\mathbf{u}}}}}

\newcommand{\vx}{{\ensuremath{{\mathbf{x}}}}}

\newcommand{\vy}{{\ensuremath{{\mathbf{y}}}}}

\newcommand{\mA}{{\ensuremath{\mathbf{A}}}}

\newcommand{\mD}{{\ensuremath{\mathbf{D}}}}

\newcommand{\mE}{{\ensuremath{\mathbf{E}}}}

\newcommand{\mH}{{\ensuremath{\mathbf{H}}}}

\newcommand{\mI}{{\ensuremath{\mathbf{I}}}}

\newcommand{\mL}{{\ensuremath{\mathbf{L}}}}

\newcommand{\mS}{{\ensuremath{\mathbf{S}}}}

\newcommand{\mU}{{\ensuremath{\mathbf{U}}}}

\newcommand{\mV}{{\ensuremath{\mathbf{V}}}}

\newcommand{\mW}{{\ensuremath{\mathbf{W}}}}

\newcommand{\mX}{{\ensuremath{\mathbf{X}}}}
\newcommand{\mY}{{\ensuremath{\mathbf{Y}}}}

\usepackage{latexsym}

\def\IC{\mathbb C}
\def\IN{\mathbb N}
\def\IZ{\mathbb Z}
\def\IR{\mathbb R}

\def\shat{^{\mathchoice{}{}%
 {\,\,\smash{\hbox{\lower4pt\hbox{$\widehat{\null}$}}}}%
 {\,\smash{\hbox{\lower3pt\hbox{$\hat{\null}$}}}}}}

\def\bSigma{{
      \ooalign{
      \smash{\hskip.4pt\raise.4pt\hbox{$\Sigma$}}\vphantom{}\crcr
      \smash{\hskip.7pt\raise.6pt\hbox{$\Sigma$}}\vphantom{}\crcr
      \smash{\hbox{$\Sigma$}}\vphantom{$\Sigma$}}
      \vphantom{\hbox{$\Sigma$}}
      }}
\def\bTheta{{
      \ooalign{
      \smash{\hskip.5pt\raise.5pt\hbox{$\Theta$}}\vphantom{}\crcr
      \smash{\hskip.0pt\raise.1pt\hbox{$\Theta$}}\vphantom{}\crcr
      \smash{\hbox{$\Theta$}}\vphantom{$\Theta$}}
      \vphantom{\hbox{$\Theta$}}
      }}
\def\bDelta{{
      \ooalign{
      \smash{\hskip.4pt\raise.4pt\hbox{$\Delta$}}\vphantom{}\crcr
      \smash{\hskip.7pt\raise.6pt\hbox{$\Delta$}}\vphantom{}\crcr
      \smash{\hbox{$\Delta$}}\vphantom{$\Delta$}}
      \vphantom{\hbox{$\Delta$}}
      }}
\def\bLambda{{
      \ooalign{
      \smash{\hskip.5pt\raise.5pt\hbox{$\Lambda$}}\vphantom{}\crcr
      \smash{\hskip.0pt\raise.1pt\hbox{$\Lambda$}}\vphantom{}\crcr
      \smash{\hbox{$\Lambda$}}\vphantom{$\Lambda$}}
      \vphantom{\hbox{$\Lambda$}}
      }}

\makeatletter

\def\bordermatrix#1{\begingroup \m@th
  \@tempdima 8.75\p@
  \setbox\z@\vbox{%
    \def\cr{\crcr\noalign{\kern2\p@\global\let\cr\endline}}%
    \ialign{$##$\hfil\kern2\p@\kern\@tempdima&\thinspace\hfil$##$\hfil
      &&\quad\hfil$##$\hfil\crcr
      \omit\strut\hfil\crcr\noalign{\kern-\baselineskip}%
      #1\crcr\omit\strut\cr}}%
  \setbox\tw@\vbox{\unvcopy\z@\global\setbox\@ne\lastbox}%
  \setbox\tw@\hbox{\unhbox\@ne\unskip\global\setbox\@ne\lastbox}%
  \setbox\tw@\hbox{$\kern\wd\@ne\kern-\@tempdima\left[\kern-\wd\@ne
    \global\setbox\@ne\vbox{\box\@ne\kern2\p@}%
    \vcenter{\kern-\ht\@ne\unvbox\z@\kern-\baselineskip}\,\right]$}%
  \null\;\vbox{\kern\ht\@ne\box\tw@}\endgroup}
\makeatother

\makeatletter
\def\argmin{\mathop{\operator@font arg\,min}}
\def\argmax{\mathop{\operator@font arg\,max}}
\makeatother

\newcommand{\bea}{\begin{array}}
\newcommand{\ena}{\end{array}}
\newcommand{\beq}{\begin{equation}}
\newcommand{\enq}{\end{equation}}

\newcommand{\beqa}{\begin{eqnarray}}
\newcommand{\enqa}{\end{eqnarray}}

\newcommand{\beqan}{\begin{eqnarray*}}
\newcommand{\enqan}{\end{eqnarray*}}

\newcommand{\AL}{\begin{enumerate}}
\newcommand{\ALE}{\end{enumerate}}

\def\addots{\mathinner{
    \mkern1mu\raise0pt\vbox{\kern7pt\hbox{.}}
    \mkern2mu\raise4pt\hbox{.}
    \mkern2mu\raise7pt\hbox{.}
    \mkern1mu}}

\def\sddots{\mathinner{
    \mkern.8mu\raise7pt\hbox{.}
    \mkern.8mu\raise4pt\hbox{.}
    \mkern.8mu\raise0pt\vbox{\kern7pt\hbox{.}}
    \mkern1mu}}

\def\saddots{\mathinner{
    \mkern.2mu\raise0pt\vbox{\kern7pt\hbox{.}}
    \mkern.2mu\raise4pt\hbox{.}
    \mkern.2mu\raise7pt\hbox{.}
    \mkern1mu}}

\def\sqplus{\mathbin{
	{\ooalign{\hfil\raise.3ex\hbox{\scriptsize
	+}\hfil\crcr\mathhexbox274\crcr\mathhexbox275}}
	}} 
\def\sqminus{\mathbin{
	{\ooalign{\hfil\raise.3ex\hbox{\scriptsize
	--}\hfil\crcr\mathhexbox274\crcr\mathhexbox275}}
	}}

\def\IC{{
   \mathord{
      \hbox to 0em{
	 \hskip-4pt
         \ooalign{
	   \smash{\hskip1.9pt\raise2.6pt\hbox{$\scriptscriptstyle |$}}\crcr
	   \smash{\hbox{\rm\sf C}} }
	 \hidewidth}
      \phantom{\hbox{\rm\sf C}}
} }}
\def\IN{
    {\ooalign{
   \smash{\hskip2.2pt\raise1.5pt\hbox{$\scriptscriptstyle |$}}\vphantom{}\crcr
   \hbox{\sf N}
	}}
	} 
\def\IZ{
    {\ooalign{
   \smash{\hskip1.9pt\raise0pt\hbox{$\sf Z$}}\vphantom{}\crcr
   \hbox{\sf Z}
	}}
	} 
\def\IR{
    {\ooalign{
   \smash{\hskip2.2pt\raise1.5pt\hbox{$\scriptscriptstyle |$}}\vphantom{}\crcr
   \smash{\hskip2.2pt\raise3.3pt\hbox{$\scriptscriptstyle |$}}\vphantom{}\crcr
   \hbox{\sf R}
	}}
	} 

\DeclareMathAlphabet{\mathcmb}{OT1}{cmr}{b}{n}

\def\bSigma{\ensuremath{\mathcmb{\Sigma}}}
\def\bLambda{\ensuremath{\mathcmb{\Lambda}}}

\def\bTheta{\ensuremath{\mathcmb{\Theta}}}

\newcommand{\SI}{\begin{indlist}}
\newcommand{\EI}{\end{indlist}}

\newcommand{\DL}{\begin{dashlist}}
\newcommand{\DLE}{\end{dashlist}}

\makeatletter
\def\setboxz@h{\setbox\z@\hbox}
\def\wdz@{\wd\z@}
\def\boxz@{\box\z@}
\def\underset#1#2{\binrel@{#2}%
  \binrel@@{\mathop{\kern\z@#2}\limits_{#1}}}
\def\binrel@#1{\begingroup
  \setboxz@h{\thinmuskip0mu
    \medmuskip\m@ne mu\thickmuskip\@ne mu
    \setbox\tw@\hbox{$#1\m@th$}\kern-\wd\tw@
    ${}#1{}\m@th$}%
  \edef\@tempa{\endgroup\let\noexpand\binrel@@
    \ifdim\wdz@<\z@ \mathbin
    \else\ifdim\wdz@>\z@ \mathrel
    \else \relax\fi\fi}%
  \@tempa
}
\let\binrel@@\relax%
\makeatother

\begin{document}

\title{Task-driven Heterophilic Graph Structure Learning
\thanks{This work is supported in part by the Kotak IISc AI–ML Center (KIAC).}
}
\author{\IEEEauthorblockN{Ayushman Raghuvanshi$^{1}$, Gonzalo Mateos$^{2}$, and Sundeep Prabhakar Chepuri$^{1}$}
\IEEEauthorblockA{\textit{$^{1}$ Indian Institute of Science, Bangalore, India. \quad $^{2}$ University of Rochester, Rochester, United States}
}}

\maketitle

\begin{abstract}
Graph neural networks (GNNs) often struggle to learn discriminative node representations for heterophilic graphs, where connected nodes tend to have dissimilar labels and feature similarity provides weak structural cues. We propose \emph{frequency-guided graph structure learning} (FgGSL), an end-to-end graph inference framework that jointly learns homophilic and heterophilic graph structures along with a spectral encoder. FgGSL employs a learnable, symmetric, feature-driven masking function to infer said complementary graphs, which are processed using pre-designed low- and high-pass graph filter banks. A label-based structural loss explicitly promotes the recovery of homophilic and heterophilic edges, enabling task-driven graph structure learning. We derive stability bounds for the structural loss and establish robustness guarantees for the filter banks under graph perturbations. Experiments on six heterophilic benchmarks demonstrate that FgGSL consistently outperforms state-of-the-art GNNs and graph rewiring methods, highlighting the benefits of combining frequency information with supervised topology inference.
\end{abstract}



\section{Introduction} \label{sec:intro} 
Graphs provide an effective framework for modeling relational patterns in complex domains, such as social networks, biochemical systems, and financial markets, where the connectivity among entities conveys valuable information about the entities themselves. Recent advances in \ac{grl} have enabled the integration of this structural information into parameterized predictive models, significantly improving performance on tasks such as node classification, link prediction, and node regression~\cite{hamilton_graph_2022}. 

Most \ac{grl}  methods, including message passing neural networks (MPNNs) and \acp{gnn}, rely on the assumption of \textit{homophily}, meaning that connected nodes tend to be semantically similar. This assumption aligns with aggregation-based and low-pass filtering approaches that promote smooth node embeddings. However, many real-world networks, such as protein--protein interaction and online discussion networks, exhibit \textit{heterophily}, where connected nodes often have dissimilar labels~\cite{pei_geom-gcn_2020}. In such cases, conventional GNNs designed under the homophilic assumption tend to perform poorly due to feature mixing between unrelated nodes~\cite{luan_heterophilic_2024}.

Research on representation learning for heterophilic graphs aims to overcome the limitations imposed by the homophily assumptions underlying standard \acp{gnn}. Approaches include redefining architectures, aggregation mechanisms, and customizing spectral filters to capture broader frequency components of graph signals \cite{abu-el-haija_mixhop_2019, bo_beyond_2021, zhu_beyond_2020, pei_geom-gcn_2020}. These methods primarily enhance the GNN’s ability to exploit graph structure and handle heterophily more effectively. 

Alternatively, \textit{graph rewiring} approaches aim to modify the graph structure to make it compatible with legacy \acp{gnn}. These methods add homophilic edges, prune heterophilic edges, or use score-based strategies to yield a more suitable topology~\cite{bi_make_2022, tenorio_adapting_2025}. However, their effectiveness is limited because most rewiring strategies rely primarily on node features to guide structural updates. In heterophilic settings, feature similarity is a poor proxy for label similarity, resulting in the introduction of edges that are not semantically meaningful, even after rewiring. As a result, many rewired graphs perform only marginally better than, or even comparably to, graph-agnostic models such as multilayer perceptrons (MLP). This first challenge highlights the need for a principled, task-driven framework that can leverage label information to infer graph structure in heterophilic settings.

\begin{figure}
    \centering
    \includegraphics[width=0.9\columnwidth]{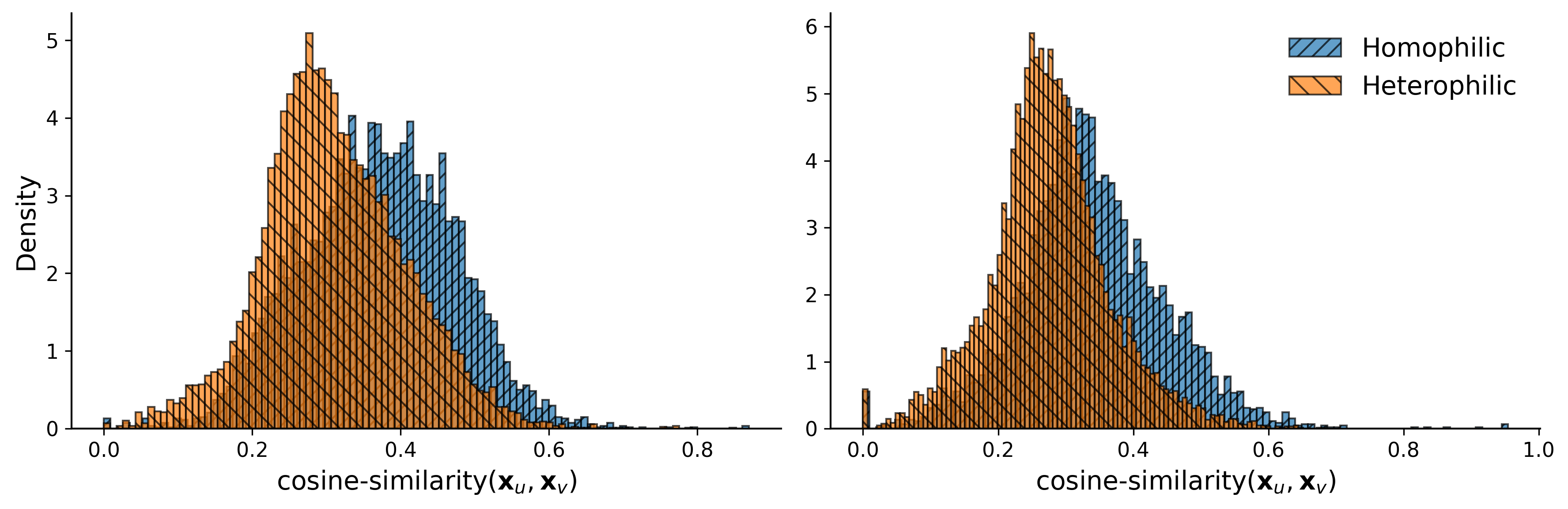}
    \caption{\footnotesize Distribution of cosine similarities of node features $\vx$ for node pairs in the \texttt{Texas} (left) and \texttt{Cornell} (right) heterophilic datasets. Homophilic pairs (same labels) and heterophilic pairs (different labels) both exhibit overlapping distributions, implying that feature similarity alone is insufficient to recover graph structures.}
    \label{fig: Feature Distribution}
    \vspace{-10pt}
\end{figure}

The goal of graph inference is to recover an underlying graph for given nodal features and subsequently apply \ac{grl} methods~\cite{dong_learning_2019}. Smoothness-based graph inference approaches attempt to identify a graph that promotes feature smoothness while enforcing structural properties such as edge sparsity. However, these techniques also face challenges on heterophilic datasets, where node feature similarity does not correlate well with label similarity. In other words, same-label and different-label node pairs present highly overlapping feature-similarity distributions, as shown in Fig.~\ref{fig: Feature Distribution}, underscoring the lack of separability between intra-class and inter-class node pairs when relying solely on node features. 
%

\vspace{2pt}\noindent\textbf{Contributions.}
To address the aforementioned twofold challenge, 
we propose the \textit{frequency-guided graph structure learning (FgGSL)} framework that jointly learns task-aware homophilic and heterophilic graph structures in a supervised setting. FgGSL learns two distinct masking functions that generate homophilic and heterophilic graphs from a fully connected graph, and then applies distinct graph filters designed to capture complementary regions of the graph spectral domain. The homophilic graph leverages pre-designed low-pass filters to smooth out the features of connected nodes, while applying node label-based smoothness regularization to promote homophilic graph recovery. On the other hand, the heterophilic graph utilizes pre-designed high-pass filters to emphasize differences in features between connected nodes, and a node label-based anti-smoothness term that guides the recovery of heterophilic graphs. By integrating the filtered signals from these two complementary graphs, FgGSL captures the rich spectral characteristics of the node features in these complex datasets.

We further establish theoretical bounds on the deviation of the filtered output for graphs close to the optimal graph structure, following the approach in \cite{gama_stability_2020}. Hence, the model remains robust even if the recovered graph is sub-optimal, as long as it is close to the optimal structure. Since testing node labels are unavailable during training, we use predicted labels and guarantee that the structural loss remains close to the actual label-based counterpart, as long as the prediction error is bounded. This enables FgGSL to maintain its performance even without the exact labels of the testing node.

To evaluate the effectiveness of the proposed framework, we perform node classification experiments on heterophilic benchmarks~\cite{pei_geom-gcn_2020}. The results demonstrate that FgGSL consistently outperforms existing \acp{gnn} designed for heterophilic settings as well as graph rewiring approaches. Moreover, the learned embedding exhibits a markedly different similarity distribution among the semantically similar and dissimilar node pairs. In summary, our work advances task-driven graph structure learning under heterophily by addressing the limitations of feature-only rewiring and (fixed-graph) \acp{gnn} via the following key contributions:
\begin{itemize}
    \item We introduce a frequency-guided, task-driven graph inference framework that jointly learns homophilic and heterophilic graph structures tailored to downstream prediction tasks;
    
    \item We propose label-aware structural loss functions that explicitly encourage homophily and heterophily, enabling supervised as well as interpretable graph inference; and
    
    \item We develop an interpretable spectral architecture that combines predefined low-pass and high-pass filter banks to capture complementary frequency components in node representations.
\end{itemize}

We empirically validate that, by jointly modeling low- and high-frequency modes through complementary spectral filters, FgGSL effectively captures informative relationships that conventional aggregation-based \acp{gnn} fail to exploit.

\section{Background}
Let  $\cG = (\cV, \cE)$ be an undirected graph, where $\cV$ is the node set with $|\cV| = N$ and $\cE$ represents the set of edges. 
Each node $i \in \cV$ is associated with a feature vector $\vx_i \in \mathbb R^F$, and the collection of all node features is represented by the matrix $\mX \in \mathbb{R}^{N \times F}$. 
One-hot encoded node labels $\vy_v$ for $v\in \cV$ are stored in $\mY  \in \{0,1\}^{ N\times C}$, where $C$ is the number of classes. 
The adjacency matrix $\mA \in \{0,1\}^{N \times N}$ encodes the graph connectivity, where $A_{ij} = 1$ if $(i, j) \in \cE$ and $A_{ij} = 0$, otherwise. 
The degree matrix is $\mD = \mathrm{diag}(\sum_j A_{ij})$, and the normalized Laplacian is given by $\mL = \mI - \mD^{-1/2}\mA\mD^{-1/2}$.
\begin{figure}
    \centering
    \includegraphics[width=0.7\columnwidth]{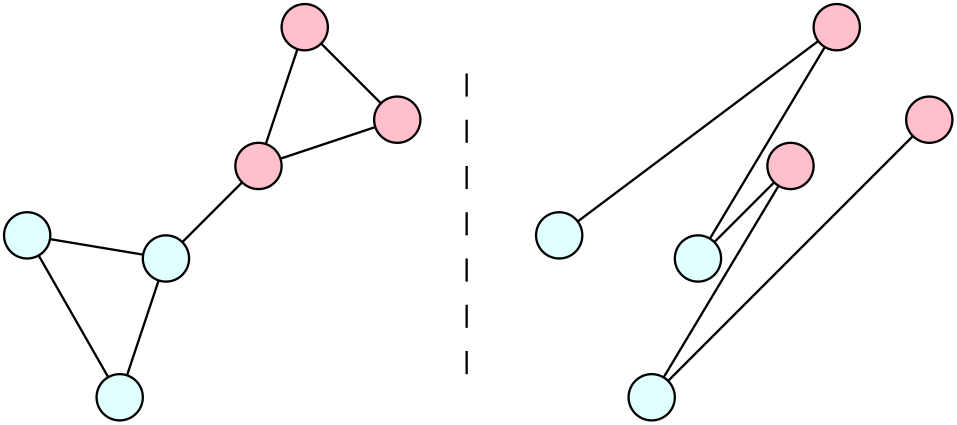}
    \caption{Illustration of a homophilic graph (left), and a heterophilic graph (right). The color of the nodes indicates their labels.}
    \vspace{-10pt}
    \label{fig:hetero_homo_graph}
\end{figure}

\vspace{2pt}\noindent\textbf{Heterophilic graphs.}
Graphs in which adjacent nodes are often semantically different, meaning they have more inter-class edges than intra-class edges, are known as \textit{heterophilic graphs} (see Fig.~\ref{fig:hetero_homo_graph}). One way to quantify heterophily is via the \emph{heterophilic edge ratio}, defined as the ratio of heterophilic edges to the total number of edges in the graph, i.e., 
$R_{\text{het}}(\cG) = |\{(u,v)\in \cE : \vy_u \neq \vy_v\}|/|\cE|.$


\vspace{2pt}\noindent\textbf{\Ac{gft}. } The \ac{gft} is an orthonormal transform that decomposes a graph signal into different “frequency” modes of graph-dependent variation. Low frequencies correspond to smooth signals (connected nodes have similar values), while high frequencies imply signals exhibit high variability across neighbors. We leverage this spectral interpretation of graph signals to explicitly extract low-frequency (smooth) signals on the learned homophilic graph and high-frequency signals on the learned heterophilic graph. 

For a graph $\cG$ whose Laplacian has eigendecomposition $\mL = \mU \boldsymbol{\Lambda} \mU^\top$, the \ac{gft} of a graph signal $\mX$ is defined as $\tilde{\mX} = \mU^\top \mX$, where $\tilde{\vx}_i^\top = \vu_i^\top \mX$ denotes the spectral component of $\mX$ associated with eigenvalue (i.e., frequency) $\lambda_i$. 
Each eigenpair $(\lambda_i, \vu_i)$ represents a frequency mode of the graph with smaller eigenvalues corresponding to smooth, low-frequency signal components. 

A \emph{graph filter} is a function of $\mL$ that can impose desired spectral characteristics onto a graph signal. It is specified in terms of a spectral kernel $h: \mathbb{R} \mapsto \mathbb{R}$, which modulates the frequency content of a signal. 
Filtering a graph signal $\mX$ with kernel $h(\cdot)$ boils down to
\begin{equation*}
    \bar{\mX} = h(\mL)\mX = \mU h(\boldsymbol{\Lambda}) \mU^\top \mX = \mU h(\boldsymbol{\Lambda}) \tilde{\mX},
\end{equation*}
where $h(\boldsymbol{\Lambda}) = \mathrm{diag}(h(\lambda_1), \ldots, h(\lambda_N))$ is a diagonal matrix containing the spectral response of the filter. 
Here, $h(\lambda_i)$ determines the degree of amplification or attenuation applied to the spectral component corresponding to frequency $\lambda_i$.

\vspace{2pt}\noindent\textbf{Graph structure inference.} The problem of graph inference can be separated into two stages: (i) unsupervised inference of a graph structure, often guided by smoothness assumptions and regularizers; and (ii) supervised training of a \ac{gnn} on the inferred graph.

Formally, given a graph signal $\mX$ over nodes $\cV$ along with labels $\mY$, the goal in (i) is to learn an optimal graph Laplacian $\mL^\star$ given a structural loss $\ell _{st}: \mathbb{R}^{N\times N} \times \mathbb{R}^{N \times f} \mapsto \mathbb{R}$ as
%
\begin{equation}\label{eqn: (i) graph learning}
    \mL^\star = \arg\min_{\mL \in \cL} , \ell_{st}(\mL, \mX),
\end{equation}
where $\cL$ is the set of valid Laplacians. Typically, for $\ell_{st}(\mL, \mX)$, we use the quadratic smoothness $\operatorname{Tr}(\mX^\top \mL \mX)$ for learning homophilic graphs or anti-smoothness $-\operatorname{Tr}(\mX^\top\mL\mX)$ for learning heterophilic graphs. This is followed by (ii), namely,
\begin{equation}\label{eqn: (ii) GNN training}
 \phi^\star = \arg\min_{\phi} \, \| \mY - f_\phi(\mL^\star, \mX) \|,   
\end{equation}
where $f_\phi$  is a \ac{gnn} with the learnable parameter $\phi$.

Solving this problem directly is challenging for heterophilic graph signals because score functions based solely on node feature similarity, such as anti-smoothness is not useful as shown in Fig.~\ref{fig: Feature Distribution}, where recall that the feature-similarity distribution of same-class node pairs overlaps with that of different-class pairs
Consequently, the learned Laplacian $\mL^\star$ constructed solely from node feature similarity does not necessarily exhibit heterophily. 


\section{Proposed Model: Frequency-guided Graph Structure Learning (FgGSL)}

\begin{figure*}
    \centering
    \includegraphics[width=1.9\columnwidth]{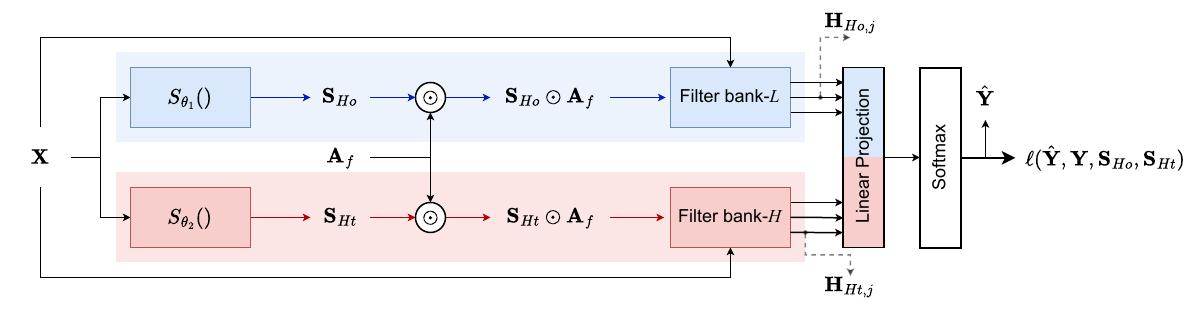}
    \caption{\footnotesize \textbf{FgGSL framework}: Node features $\mX$ generate two parameterized graph structures by masking a fully connected adjacency matrix $\mA_f$ with learnable functions $S_{\theta_1}$ and $S_{\theta_2}$. The homophilic graph $\mS_{Ho} \odot \mA_f$ uses a low-pass filter bank, while the heterophilic graph $\mS_{Ht} \odot \mA_f$ employs a high-pass filter bank. Filter outputs $\mH_{Ho,j}$ and $\mH_{Ht,j}$ at different scales are concatenated and passed through a linear layer with softmax to produce class probabilities $\hat{\mY}$.}
    \label{fig:Model-Arch}
    \vspace{-5pt}
\end{figure*}

In this section, we reformulate \eqref{eqn: (i) graph learning} and \eqref{eqn: (ii) GNN training} as a joint graph learning and encoder training problem. We introduce two label-based structural loss functions that incorporate label similarity into the structural score. We then develop a parameterized method to explore the graph search space, define filter banks to generate node representations, and introduce a label-based structural loss that guides the learning of strong heterophilic and homophilic graphs. We judiciously combine these ingredients to arrive at the novel FgGSL framework. 

\vspace{2pt}\noindent\textbf{Problem formulation.} 
Consider a heterophilic graph $\cG$ with node features $\mX$ and one-hot encoded class labels $\mY$. 
Our goal is to jointly learn an homophilic graph $\hat{\cG}_{\text{Ho}}$, a heterophilic graph $\hat{\cG}_{\text{Ht}}$, 
and the parameters $\hat\phi$ of an encoder $f_{{\phi}}$ by solving the optimization problem:
\begin{equation} \label{eq:total_loss}
    \{\hat{\cG}_{\text{Ho}}, \hat{\cG}_{\text{Ht}}, \hat{\phi}\} 
    = \underset{\cG_{\text{Ho}},\,\cG_{\text{Ht}},\,\phi}{\operatorname{argmin}}  \; \ell (\cG_{\text{Ho}}, \cG_{\text{Ht}}, \mX, f_\phi ,\mY). 
\end{equation}
Here, $\ell$ is the joint graph structure and encoder training loss function that takes the form 
\begin{align*}
    \ell={}& \ell_{\text{CE}}\big(f_{\phi}(\cG_{\text{Ho}}, \cG_{\text{Ht}}, \mX), \mY\big)
    + \ell_{\text{SL}}(\cG_{\text{Ho}},\,\cG_{\text{Ht}},\,\mY),\\
    \ell_{\text{SL}} 
    ={}& \alpha\,\ell_{\text{Ho}}(\cG_{\text{Ho}}, \mY)
    + \beta\,\ell_{\text{Ht}}(\cG_{\text{Ht}}, \mY),
\end{align*}
where $\alpha, \beta > 0$ are hyper-parameters. 
The cross-entropy loss $\ell_{\text{CE}}$ is the task-driven loss used to train the encoder; alternatively, other losses may be used depending on the downstream task. The second term $\ell_{\text{SL}}$ is the structural loss that includes both a homophilic loss $\ell_{\text{Ho}}$, promoting label smoothness over $\cG_{\text{Ho}}$, 
and a heterophilic loss $\ell_{\text{Ht}}$, promoting label dissimilarity over $\cG_{\text{Ht}}$; see also \eqref{eq:struct_loss_ho}-\eqref{eq:struct_loss_ht} for details. The FgGSL framework, illustrated in Fig.~\ref{fig:Model-Arch}, minimizes the loss by first parametrizing the graph structure via learnable masking, then applying spectral filtering using filter banks. The outputs of the filters are concatenated and linearly combined to produce the logits for node classification

\vspace{2pt}\noindent\textbf{Parameterized graph structure.} 
To make the graph structure learnable, we express $\cG_{\text{Ho}}$ and $\cG_{\text{Ht}}$ as masked versions of a fully connected graph $\cG_{f} = (\cV, \cE_{f})$ with adjacency $\mA_f$.
This parameterization enables the model to adjust edge weights and effectively infer the optimal connectivity pattern based on node features.
Specifically, we define a learnable masking function $S_\theta : \mathbb{R}^F \times \mathbb{R}^F \mapsto [0,1]$ that assigns a weight $w_{ij} \in [0,1]$ to each edge $(i,j)$ in $\cE_{f}$.
This masking function is defined as the inner product between transformed node features. Therefore, for a parameterized nonlinear map $\Phi_\theta: \mathbb{R}^F \mapsto \mathbb{R}^D$, the mask is computed as:
\begin{equation}\label{eqn: mask}
S_\theta(\vx_u, \vx_v) = \sigma\big(\Phi_\theta^\top(\vx_u), \Phi_\theta(\vx_v)\big) = w_{uv},\quad (u,v)\in\cE_{f},
\end{equation}
where $\sigma(\cdot)$ denotes the sigmoid activation.
Thus, for a given graph $\cG_{f}$, we obtain a masked homophilic graph $\hat{\cG}_{\text{Ho}} = (\cV, \cE_{f}, S_{\hat{\theta}1})$ and a heterophilic graph $\hat{\cG}_{\text{Ht}} = (\cV, \cE_{f}, S_{\hat{\theta}2})$, with adjacencies $\mS_{\text{Ho}}\odot\mA_f$ and $\mS_{\text{Ht}}\odot\mA_f$, respectively.  Therefore, the search space for the graph is parameterized by the learnable weights $\theta$ in the nonlinear map $\Phi_\theta$, which govern the edge masking process. In particular, we use parameters $\theta_1$ for the homophilic graph and $\theta_2$ for the heterophilic graph. Intuitively, edges with weights close to zero contribute minimally to the learned structure. 
When $\cG_{f}$ is fully connected, this formulation allows the model to infer any possible graph as a masked variant of the initial one. 
Alternatively, when $\cG_{f}$ corresponds to some graph available in the dataset, the masking operation effectively performs edge pruning. In practice, $\cG_{f}$ may be either fully connected or the given graph.

\vspace{2pt}\noindent\textbf{Filter banks.} 
The function $f_{\phi}$, optimized in~\eqref{eq:total_loss}, represents the classifier that produces class probabilities $\hat{\mY}\in [0,1]^{ N\times C}$ from node features $\mX$ and the learned graphs $\{\hat{\cG}_{\text{Ho}},\hat{\cG}_{\text{Ht}}\}$. 
In FgGSL, the classifier employs two sets of graph filters: low-pass filters applied to the homophilic graph $\cG_{\text{Ho}}$ and high-pass filters applied to the heterophilic graph $\cG_{\text{Ht}}$. 
The low-pass filters extract smooth, shared information among similar nodes, while the high-pass filters highlight discriminative or contrasting information among dissimilar nodes.

\begin{figure}
    \centering
    \includegraphics[width=0.8\columnwidth]{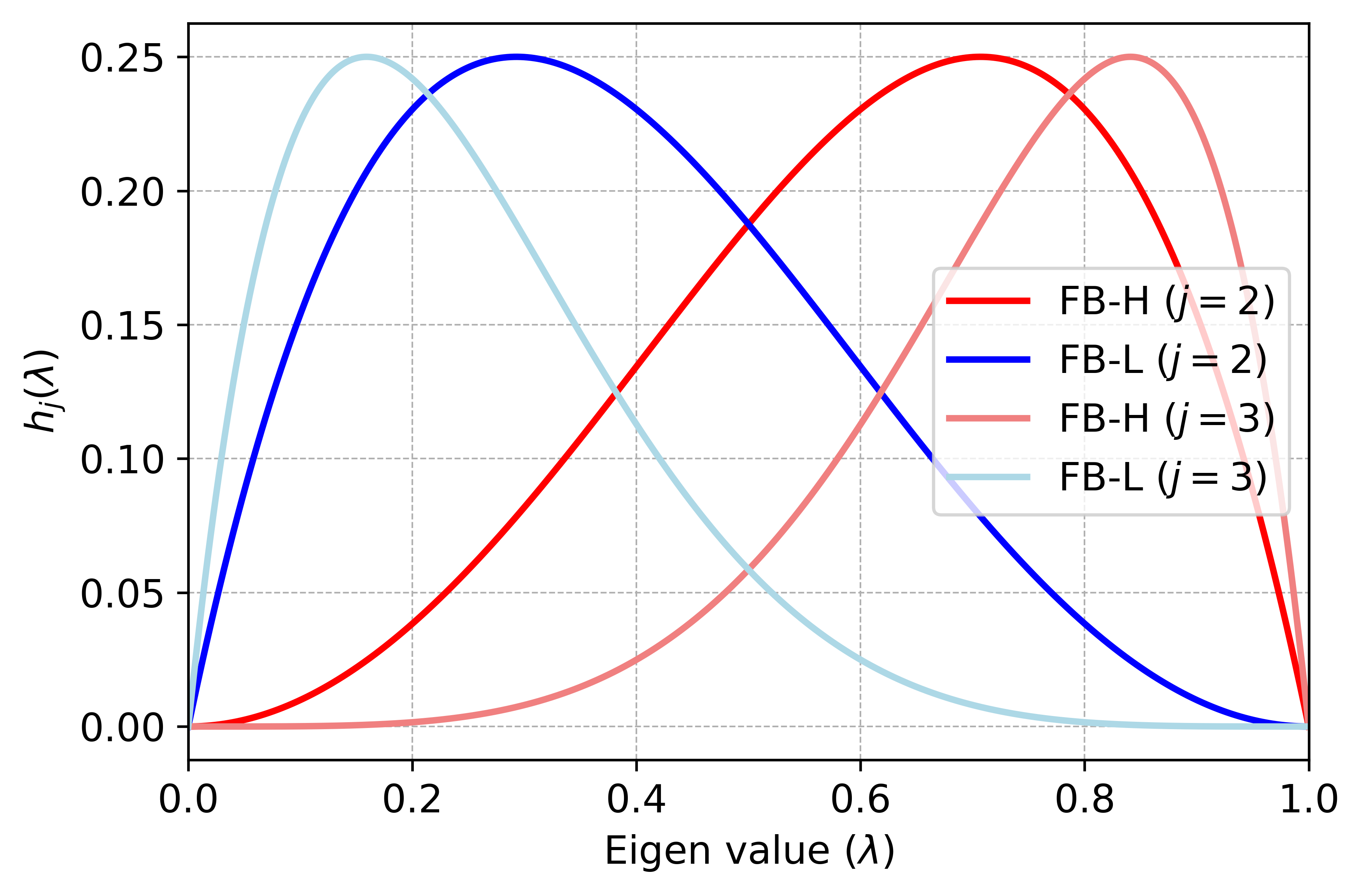}
    \caption{\footnotesize Filter-bank frequency responses: Filters from filter bank-$L$ preserve information from the lower end of the spectrum (blue), while filters from filter bank-$H$ preserve information from the higher end of the spectrum (red).}
    \label{fig:filter_bank}
    \vspace{-10pt}
\end{figure}

Specifically, we use predefined low-pass filter banks $\{h_L^{(j)}\}_{j=2}^J$ on $\cG_{\text{Ho}}$ and high-pass filter banks $\{h_H^{(j)}\}_{j=2}^J$ on $\cG_{\text{Ht}}$. For a given maximum scale $J$, the filter kernels are defined as
\begin{align}\label{eqn: filter-banks}
    h_L^{(j)}(\lambda) &= \left(0.5\lambda\right)^{2^{j-1}} - \left(0.5\right)^{2^{j}}, \\
    h_H^{(j)}(\lambda) &= \left(1 - 0.5\lambda\right)^{2^{j-1}} - \left(1 - 0.5\lambda\right)^{2^{j}}
\end{align}
These polynomial \textit{diffusion filters} are illustrated in Fig.~\ref{fig:filter_bank} at different scales. They admit efficient implementation through repeated Laplacian multiplications, eliminating the need for explicit eigendecomposition.  Given the input node features $\mX$, the model aggregates the responses of all filters from both graphs as $\mH = [\,\mH_L \mid \mH_H\,] \in \mathbb{R}^{N \times 2(J-1)F}$, where
\begin{align*}
    \mH_L &= [\,h_L^{(2)}(\mL_{\text{Ho}})\mX \mid h_L^{(3)}(\mL_{\text{Ho}})\mX \mid \cdots \mid h_L^{(J)}(\mL_{\text{Ho}})\mX\,], \\
    \mH_H &= [\,h_H^{(2)}(\mL_{\text{Ht}})\mX \mid h_H^{(3)}(\mL_{\text{Ht}})\mX \mid \cdots \mid h_H^{(J)}(\mL_{\text{Ht}})\mX\,].
\end{align*}
Finally, a linear layer followed by a softmax function produces node-level class probabilities: $\hat{\mY} = \operatorname{softmax}(\mH \mW_\phi)$, for $\mW_\phi \in \mathbb{R}^{2(j-1)F \times C}.$
Here, \(\phi\) denotes the learnable weights of the linear layer in the classifier \(f_\phi\).

\vspace{2pt}\noindent\textbf{Structural loss.} 
To explicitly enforce homophily and heterophily in the inferred graph, we define two \emph{label-based structural losses}. 
The \emph{homophilic structural loss} $\ell_{\text{Ho}}$ penalizes edges connecting dissimilar labels, while the \emph{heterophilic structural loss}  $\ell_{\text{Ht}}$ penalizes edges connecting similar labels. Formally, they are defined as:
\begin{align}
    \ell_{\text{Ho}}(\cG_{\text{Ho}}, \mY) &= \sum_{(i,j) \in \cE_f} w^{(1)}_{ij} \, (1-\operatorname{cosine}(\vy_i, \vy_j)),\label{eq:struct_loss_ho} \\
    \ell_{\text{Ht}}(\cG_{\text{Ht}}, \mY) &= \sum_{(i,j) \in \cE_f} w^{(2)}_{ij} \, \operatorname{cosine}(\vy_i, \vy_j),\label{eq:struct_loss_ht}
\end{align}
where $\operatorname{cosine}(\vy_i,\vy_j) = \vy_i^\top \vy_j/(\|\vy_i\|_2 \, \|\vy_j\|_2)$ is large for similar label vectors and $1-\operatorname{cosine}(\vy_i,\vy_j)$ is large for dissimilar ones. This ensures that a larger weight penalty is applied to the edge weights of nodes with the same class in the heterophilic mask, and a larger penalty is applied to the edge weights of nodes with different classes in the homophilic mask, thereby enforcing the heterophilic and homophilic structures for $\cG_{\text{Ht}}$ and $\cG_{\text{Ho}}$, respectively.
Recall the masking function introduced in \eqref{eqn: mask}, which outputs the  weight $w^{(k)}_{ij} = \sigma(\Phi_{\theta_k}^\top(\vx_i) \Phi_{\theta_k}(\vx_j))$ for edge $(i,j)\in\cE_f$.  
Since true labels are only available for a subset of nodes in the training set, we use the predicted softmax probabilities $\hat{\mY}$ in the structural loss. 
As we argue next in Proposition~\ref{prop:structural}, the deviation between true and predicted label similarities is bounded by the prediction error, so the structural loss approaches the ideal label-based objective as predictions improve. Thus, the total loss $\ell$ in \eqref{eq:total_loss} is approximated as
\begin{equation*}
    \ell = \ell_{\text{CE}}(\hat{\mY}_{\text{train}},\mY_{\text{train}})
    + \ell_{\text{SL}}(\cG_{\text{Ho}},\,\cG_{\text{Ht}},\,\hat{\mY}),
\end{equation*}
where $\ell_{\text{SL}}$ follows the formulation in~\eqref{eq:total_loss} with $\mY$ replaced by $\hat{\mY}$. $\ell$ is minimized with respect to the learnable parameters $\theta_1$ and $\theta_2$ of the two masks, as well as the weights of the linear layer $\phi$.


\section{Theoretical and Complexity Analyses}

Here, we derive stability guarantees to justify replacing labels with predictions and show that the filter bank output exhibits robustness to graph perturbations. Moreover, we discuss the complexity of FgGSL.

\vspace{2pt}\noindent\textbf{Structural loss stability.} The structural loss $\ell_{\text{SL}}$ is computed using the predicted labels \(\hat{\mY}\) in lieu of the ground-truth labels \(\mY\). Using Proposition \ref{prop:structural} below, we justify that this surrogate introduces only a controlled error, bounded by the discrepancy between \(\mY\) and \(\hat{\mY}\).
\begin{Proposition}
\label{prop:structural}
Let $C$ be the number of classes. 
%
Let $\hat{\vy}_i$ and $\hat{\vy}_j$ be the predicted class probabilities of the node vectors, and suppose $\|\vy_i - \hat{\vy}_i\|_2 \le \epsilon_i$. Then
$$
\big|
\operatorname{cosine}(\vy_i,\vy_j)
-
\operatorname{cosine}(\hat{\vy}_i,\hat{\vy}_j)
\big|
\;\le\;
2\sqrt{C} \left(
\epsilon_i
+
\epsilon_j
\right).
$$
\end{Proposition}
\noindent The proof follows by writing the difference of the cosine similarities as an inner product of the unit vectors and then applying the Cauchy–Schwarz inequality.

\noindent Thus, as the classifier accuracy improves, the adopted structural loss approximation comes increasingly closer to the true label-based structural loss.

\vspace{2pt}\noindent\textbf{Filter bank stability under graph perturbations.} Let $\mL^\star$ denote the optimal graph Laplacian, and let $\hat{\mL}$ be the recovered (possibly suboptimal) Laplacian used in the FgGSL filter bank. Following~\cite{gama_stability_2020}, we assume that the \emph{operator distance modulo permutation} $\|\cdot\|_\cP$ satisfies
\[
\|\mL^\star - \hat{\mL}\|_{\mathcal{P}} = \|\mE\| \le \epsilon,
\]
where $\mE$ is the absolute perturbation modulo permutation.

\noindent For a Lipschitz graph filter $h_j(\cdot)$ with Lipschitz constant \(K_h\),~\cite[Theorem 1]{gama_stability_2020} establishes the following bound on the filter outputs:
\[
\|h_j(\mL^\star) - h_j(\hat{\mL})\|_{\mathcal{P}}
\le
2K_h\,(1 + \delta \sqrt{N})\,\epsilon
+ \mathcal{O}(\epsilon^2),
\]
where $\delta := (\|\mU - \mV\|_2 + 1)^2 - 1$ quantifies the eigenvector misalignment between the Laplacian $\mL^\star=\mU \boldsymbol{\Lambda} \mU^\top$ and the error matrix $\mE = \mV \Lambda_{\mE} \mV^\top$.

\noindent Applying this result to our multiscale filter bank in \eqref{eqn: filter-banks} yields 
\[
\| h_j(\mL^\star) - h_j(\hat{\mL}) \|_{\mathcal{P}}
\le 
2^{(j-1)}(1 + \delta\sqrt{N})\,\epsilon
+ \mathcal{O}(\epsilon^2).
\]
This indicates that the filter banks remain stable under small structural perturbations of the graph, and a near-optimal graph structure will have a minimal effect on the node embeddings generated by FgGSL.

\vspace{2pt}\noindent\textbf{Computational complexity.} 
FgGSL consists of two major components: (i) mask construction; and (ii) graph filtering. 
Given a graph with $N$ nodes, the initial fully–connected candidate graph contains up to $\frac{N(N-1)}{2}$ edges. 
Computing the masking weights defined as inner products of transformed node features requires $\mathcal{O}(N^2)$ operations. After the masked graph is obtained, applying a single filter of the form $h_j(\mathbf{L})\mathbf{X}$ incurs a cost of $\mathcal{O}(N^2 f)$, where $f$ is the feature dimension. 
Constructing the set of $2$–to–$J$ filters, i.e., $[\,h_2(\mathbf{L}), h_3(\mathbf{L}), \ldots, h_J(\mathbf{L})\,]$, requires $\mathcal{O}(J N^3)$ time due to repeated multiplication with the graph Laplacian. 
Overall, the dominant cost arises from building the multi-scale filter bank, yielding a total complexity of $\mathcal{O}(J N^3 + N^2 f)$.

\begin{table*}[!t]
\centering
\caption{Performance Comparison on Heterophilic Graph Datasets.}
\label{tab: results}
\begin{tabular}{l|cccccc}
\textbf{Model} & 
\textbf{Texas (0.88)} & 
\textbf{Wisconsin (0.79)} & 
\textbf{Cornell (0.87)} & 
\textbf{Squirrel (0.77)} & 
\textbf{Actor (0.78)} & 
\textbf{Chameleon (0.76)} \\
\toprule
\toprule
FgGSL & 
\textbf{0.94 $\pm$ 0.08} & 
\textbf{0.96 $\pm$ 0.05} & 
\textbf{0.94 $\pm$ 0.08} & 
0.58 $\pm$ 0.09 & 
\textbf{0.41 $\pm$ 0.02} & 
\textbf{0.79 $\pm$ 0.09} \\

SAGE & 
0.74 $\pm$ 0.08 & 
0.74 $\pm$ 0.08 & 
0.69 $\pm$ 0.05 & 
0.37 $\pm$ 0.02 & 
0.34 $\pm$ 0.01 & 
0.50 $\pm$ 0.01 \\

GAT & 
0.52 $\pm$ 0.06 & 
0.49 $\pm$ 0.04 & 
0.61 $\pm$ 0.05 & 
0.40 $\pm$ 0.01 & 
0.27 $\pm$ 0.01 & 
0.60 $\pm$ 0.02 \\

MLP & 
0.79 $\pm$ 0.04 & 
0.85 $\pm$ 0.03 & 
0.75 $\pm$ 0.02 & 
0.35 $\pm$ 0.02 & 
0.35 $\pm$ 0.01 & 
0.50 $\pm$ 0.02 \\

H2GCN & 
0.80 $\pm$ 0.05 & 
0.84 $\pm$ 0.05 & 
0.70 $\pm$ 0.05 & 
\textbf{0.59 $\pm$ 0.01} & 
0.35 $\pm$ 0.01 & 
0.69 $\pm$ 0.01 \\

Geom-GNN & 
0.78 $\pm$ 0.07 & 
0.80 $\pm$ 0.06 & 
0.61 $\pm$ 0.08 & 
0.56 $\pm$ 0.02 & 
0.35 $\pm$ 0.01 & 
0.65 $\pm$ 0.02 \\

MixHop & 0.81 $\pm$ 0.09 & 0.83 $\pm$ 0.08 & 0.78 $\pm$ 0.09  &  0.35 $\pm$ 0.03 & 0.34 $\pm$ 0.01  & 0.53 $\pm$ 0.02   \\

SG-GCN & 0.83 $\pm$ 0.01 & 0.83 $\pm$ 0.01 & 0.72 $\pm$ 0.01 & 0.60 $\pm$ 0.02 & 0.36 $\pm$ 0.01 & 0.67 $\pm$ 0.03  \\

FAGCN & 0.83 $\pm$ 0.01 & 0.82 $\pm$ 0.01 & 0.71 $\pm$ 0.01 & 0.31 $\pm$ 0.02 & 0.35 $\pm$ 0.01 & 0.46 $\pm$ 0.03  \\

\hline
\end{tabular}
\end{table*}

\begin{figure}[b]
    \centering
    \includegraphics[width=0.9\columnwidth]{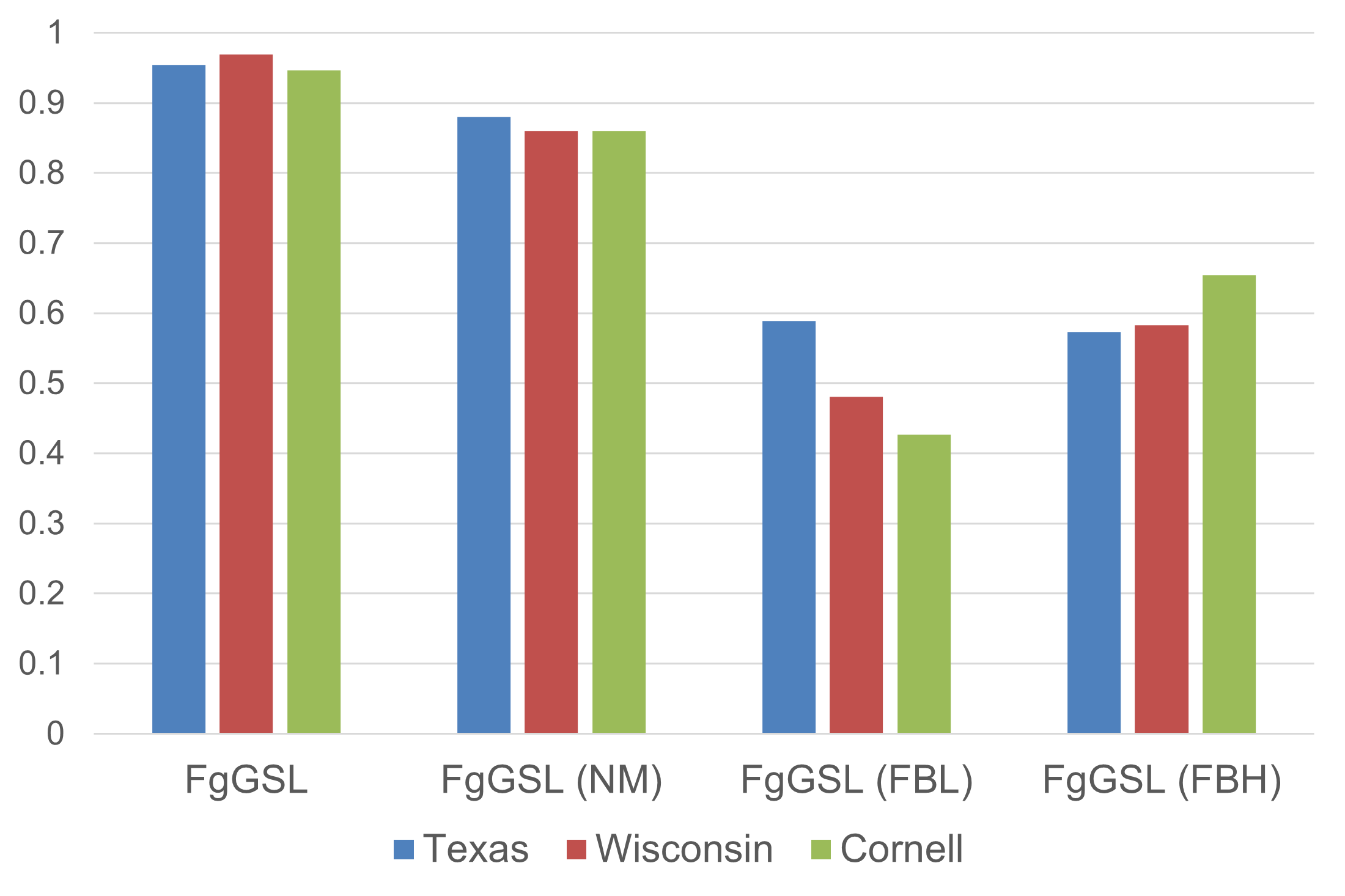}
    \caption{\footnotesize Mean accuracy of different variants of the model.}
    \label{fig:ablation}
\end{figure}
\begin{figure}
    \centering
    \includegraphics[width=\linewidth]{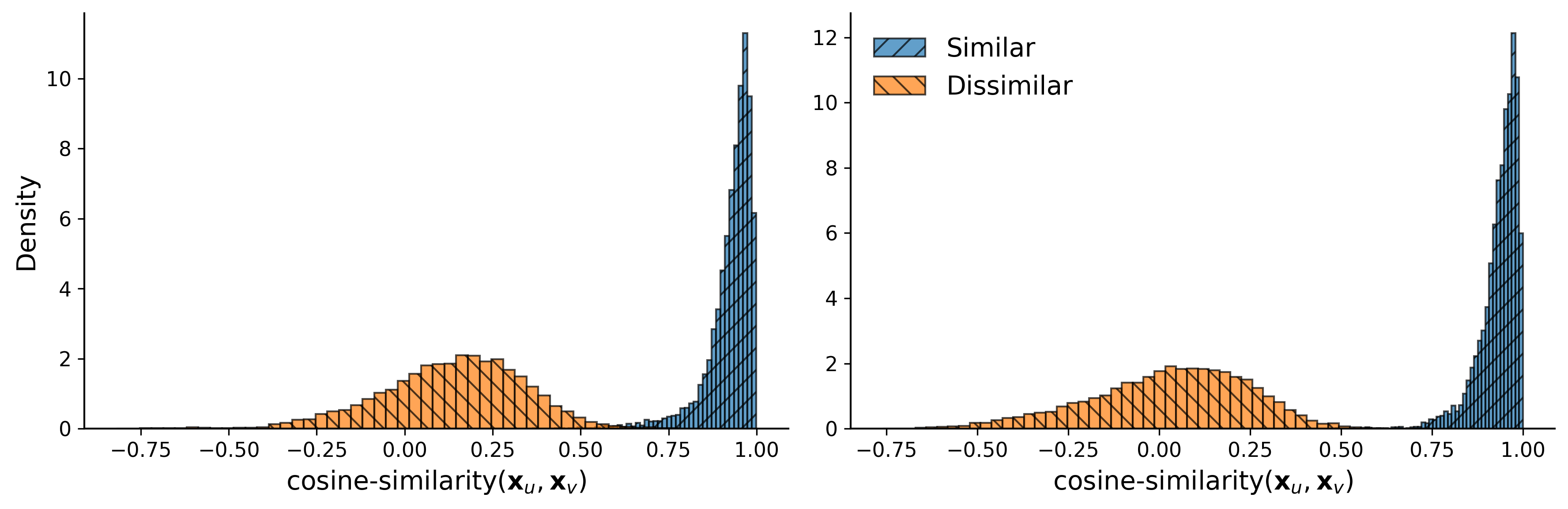}
    \caption{ \footnotesize Distribution of cosine similarities of node embeddings for all node pairs in the \texttt{Texas} (left) and \texttt{Cornell} (right) heterophilic datasets. Notice how node pairs with the same labels have higher cosine similarity compared to those with different labels, demonstrating the effectiveness of the embeddings. This is to be compared with the overlapping distributions in Fig. \ref{fig: Feature Distribution}.}
    \label{fig: embedding similarity}
    \vspace{-10pt}
\end{figure}
\section{Numerical experiments}

In this section, we evaluate the effectiveness of the proposed \texttt{FgGSL} framework on node classification tasks over 6 heterophilic graph datasets. We compare our approach against a comprehensive set of baselines: Geom-GCN \cite{pei_geom-gcn_2020}, a geometry-aware architecture that preserves structural and positional relations between nodes; GAT \cite{velickovic_graph_2018}, an attention-based GNN that learns edge-wise importance weights; GraphSAGE \cite{hamilton_inductive_2018}, an inductive neighborhood-sampling model with feature aggregation; a graph-agnostic MLP; H2GCN \cite{zhu_beyond_2020}, a heterophily-aware model that separates ego and neighbor information while leveraging higher-order neighborhoods; FAGCN \cite{bo_beyond_2021}, which incorporates high-frequency information in GCN layers; MixHop \cite{abu-el-haija_mixhop_2019}, which aggregates and mixes multi-hop neighborhood information in parallel; and SG-GCN \cite{tenorio_adapting_2025}, which generates multiple parallel graph structures based on different score functions and trains separate \acp{gnn} to produce the final node embeddings. The performance metrics are reported as mean and standard deviation of node classification accuracy over ten standard train–val–test splits. 

We further conduct an ablation study to assess the contribution of each module in our framework. Three variants are considered: $\texttt{FgGSL(NM)}$, $\texttt{FgGSL(FBL)}$, and $\texttt{FgGSL(FBH)}$. The $\texttt{FgGSL(NM)}$ variant removes the masking mechanism and applies the two parallel filter banks directly on the given graph. The $\texttt{FgGSL(FBL)}$ variant retains the masking function but employs only the low-pass filter bank to process the nodal features. Conversely, $\texttt{FgGSL(FBH)}$ uses masking followed by the high-pass filter bank only. This way, we isolate the masking mechanism and spectral filtering components. 

\vspace{2pt}\noindent\textbf{Datasets. } We conduct experiments on six widely used heterophilic benchmark datasets~\cite{pei_geom-gcn_2020}: \texttt{Texas} (heterophily ratio: 0.88), \texttt{Wisconsin} (0.79), \texttt{Cornell} (0.87), \texttt{Squirrel} (0.77), \texttt{Chameleon} (0.76), and \texttt{Actor} (0.78). The first three datasets originate from the WebKB domain and comprise webpage graphs with strong label dissimilarity. In contrast, Squirrel and Chameleon represent large Wikipedia page networks characterized by high feature dimensionality and noisy inter-class edges. The Actor dataset contains co-occurrence relations between actors in film contexts, exhibiting pronounced heterophily.

\vspace{2pt}\noindent\textbf{Results.} Across all six heterophilic graph datasets, \texttt{FgGSL} consistently achieves top or competitive accuracy. As shown in Table~\ref{tab: results}, \texttt{FgGSL} outperforms or matches the performance of existing baselines on every dataset. An ablation study, shown in Fig.~\ref{fig:ablation}, demonstrates that the full model performs best across multiple datasets. The accuracy drop in \texttt{FgGSL(NM)} highlights the importance of feature-guided masking, while the weaker performance of the single-filter variants underscores the benefit of combining both filter banks.

To further evaluate the quality of the learned node embeddings, we analyze the cosine similarity distributions of intra-class and inter-class node pairs. Unlike raw features (cf. Fig.~\ref{fig: Feature Distribution}), the embeddings from \texttt{FgGSL} exhibit a clear separation between the two types of node pairs, as illustrated in Fig.~\ref{fig: embedding similarity}. This separation demonstrates the model's effectiveness in capturing heterophilic structures.


\section{Conclusions}
In this work, we introduced FgGSL, a frequency-guided framework for joint graph structure learning and node classification on heterophilic graphs. By learning feature-driven masks for homophilic and heterophilic graphs (thereby learning two graphs) and combining low-pass and high-pass spectral filters, FgGSL recovers both homophilic and heterophilic relations that are informative for the 
downstream task. Our label-based structural loss enables supervised, task-driven graph inference and is supported by stability guarantees for both the loss and the filter bank. Empirical evaluations across six heterophilic benchmarks show that FgGSL consistently outperforms existing GNNs and rewiring approaches, while maintaining an 
interpretable and theoretically grounded design.

\bibliographystyle{IEEEtran}
\bibliography{references}

\end{document}